\pdfoutput=1

\documentclass[11pt]{article}

\usepackage{colortbl}

\usepackage{ACL2023}

\usepackage{times}
\usepackage{latexsym}

\usepackage[T1]{fontenc}
\usepackage[utf8]{inputenc}
\usepackage{microtype}
\usepackage{inconsolata}

\usepackage{bm}
\usepackage{amsfonts}
\usepackage{amsmath}
\usepackage{amssymb}
\usepackage{enumitem}
\usepackage{booktabs}
\usepackage{graphicx}
\usepackage{multirow}
\usepackage{relsize}

\title{Gradient-based Intra-attention Pruning on Pre-trained Language Models}

\author{
Ziqing Yang$^\dag$,
Yiming Cui$^\ddag$$^\dag$,
Xin Yao$^\dag$,
Shijin Wang$^\dag$$^\S$ \\
{$^\dag$State Key Laboratory of Cognitive Intelligence, iFLYTEK Research, Beijing, China} \\
{$^\ddag$Research Center for SCIR, Harbin Institute of Technology, Harbin, China} \\
{$^\S$iFLYTEK AI Research (Central China), Wuhan, China} \\
$^\dag$\tt\{zqyang5,ymcui,xinyao10,sjwang3\}@iflytek.com \\
$^\ddag$\tt ymcui@ir.hit.edu.cn}

\begin{document}
\maketitle
\begin{abstract}
Pre-trained language models achieve superior performance but are computationally expensive. Techniques such as pruning and knowledge distillation have been developed to reduce their sizes and latencies. In this work, we propose a structured pruning method \textbf{GRAIN} (\textbf{Gra}dient-based \textbf{In}tra-attention pruning), which performs task-specific pruning with knowledge distillation and yields highly effective models. Different from common approaches that prune each attention head as a whole, GRAIN inspects and prunes intra-attention structures, which greatly expands the structure search space and enables more flexible models. We also propose a gradient separation strategy that reduces the interference of distillation on pruning for a better combination of the two approaches. Experiments on GLUE, SQuAD, and CoNLL 2003 show that GRAIN notably outperforms other methods, especially in the high sparsity regime, and achieves $6\sim7\times$ speedups while maintaining $93\%\sim99\%$ performance. Under extreme compression where only $3\%$ transformer weights remain, the pruned model is still competitive compared to larger models.\footnote{Code is available at \url{https://github.com/airaria/GRAIN}.}
\end{abstract}

\section{Introduction}

Transformer-based \cite{DBLP:conf/nips/VaswaniSPUJGKP17} pre-trained language models (PLMs) have achieved great success and become the backbones of various natural language processing tasks. However, PLMs are computationally expensive and slow in inference due to their large sizes, which limits their applications in real-world scenarios. Hence, a growing interest has been in developing compression and acceleration methodologies for PLMs. 

A common approach to model compression is structured pruning, which compresses the model by removing groups of consecutive parameters, namely the pruning units. In applying structured pruning on PLMs, recent works have investigated removing units such as hidden dimensions in feed-forward layers, attention heads in the multi-head attention \cite{DBLP:conf/nips/MichelLN19,li-etal-2022-probing}, and coarse-grained units such as multi-head attention layers and feed-forward layers \cite{xia-etal-2022-structured}.

\begin{figure}[t!]
\center
  \includegraphics[width=\linewidth]{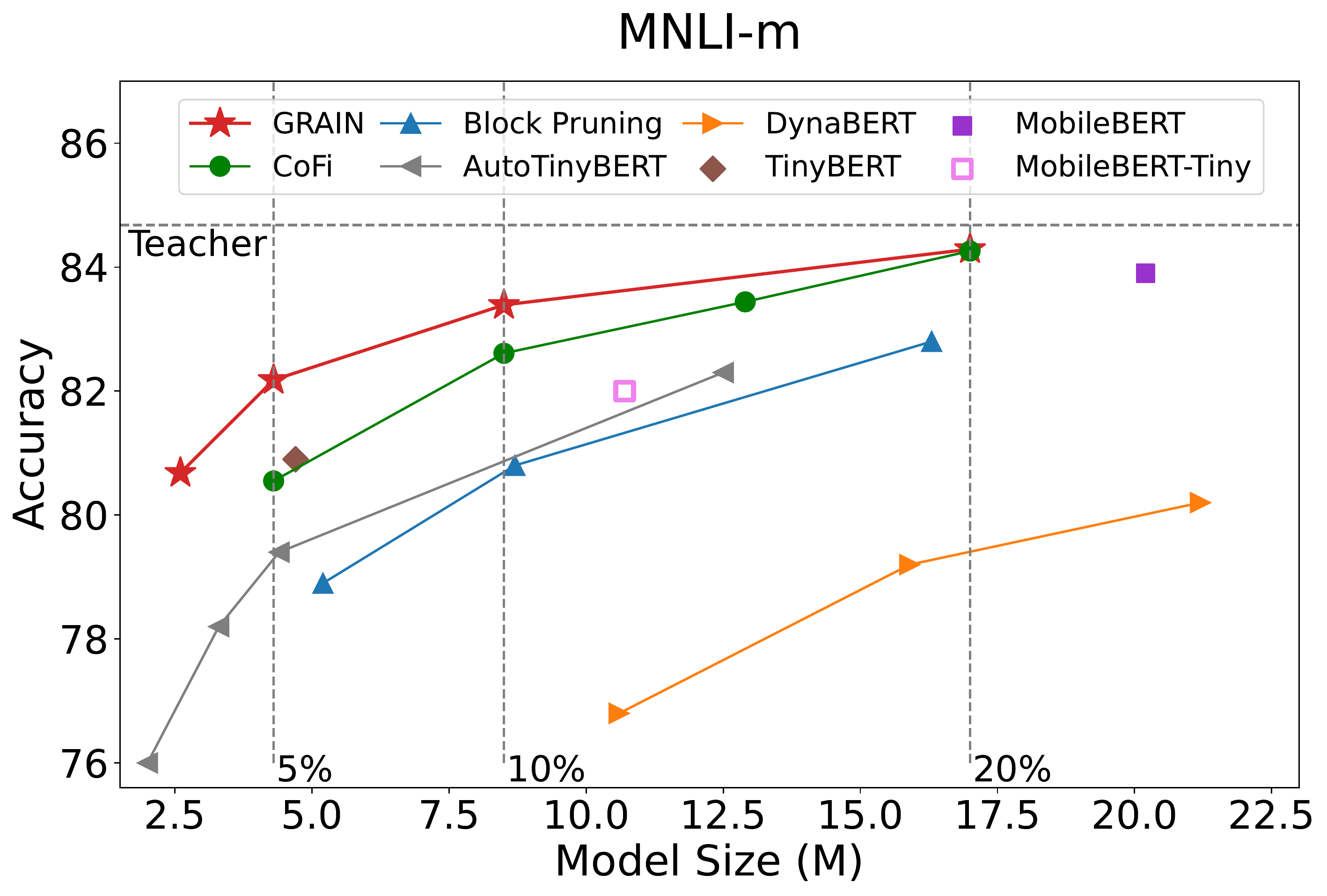} 
  \caption{A comparison of GRAIN and other distillation and pruning methods on MNLI-m development set at different model sizes. More details are in Section \ref{section:Experiments}.}
  \label{figure:mnli}
\end{figure}

However, these pruning units only span a small space of model structures and limit the exploration for better structures. For example, 
in the pruning of BERT$_{\textrm{base}}$ \cite{devlin-etal-2019-bert}, which contains $144$ attention heads, the possible choices of attention heads for the pruned model are limited.
Block Pruning \cite{lagunas-etal-2021-block} extends pruning units by considering blocks in the weight matrices, but Block Pruning is not a fully structured pruning method and can not achieve large speedups.

In this work, we propose \textbf{GRAIN} (\textbf{Gra}dient-based \textbf{In}tra-attention pruning), a structured pruning method that prunes PLMs with finer pruning units. In the following, we present the method from three aspects: pruning units, pruning algorithm, and training objectives.

\noindent\textbf{Pruning Units}\quad Unlike attention heads pruning where the pruning unit is a single head, we propose intra-attention pruning, which inspects and prunes the structures inside attention heads. Intra-attention pruning greatly expands the search space of model structures, making the resulting models more likely to find better structures. However, directly applying intra-attention pruning yields fragmented models, i.e., models with many small heads. The fragmented models have relatively large latencies on devices like GPUs. To overcome the shortcoming, we introduce structure regularization, which encourages prioritizing specific units for pruning. Structure regularization helps generate more regular structures and achieve lower latencies.

\noindent\textbf{Pruning Algorithm}\quad Pruning algorithms decide which units to be removed. We adapt the gradient-based pruning algorithm \cite{DBLP:conf/nips/MichelLN19} for intra-attention pruning. Gradient-based pruning is a light-weighted method that estimates the importance of the pruning units with gradient-based scores and then prunes the least important ones. In addition, we conduct the pruning in an iterative manner \cite{DBLP:conf/iclr/ZhuG18}, i.e., the model is gradually pruned during fine-tuning. The iterative approach has been employed in combination with pruning algorithms such as Movement Pruning \cite{DBLP:conf/nips/Sanh0R20} and Magnitude Pruning \cite{DBLP:conf/iclr/ZhuG18}, but few works have combined it with gradient-based pruning. We find that iterative gradient-based pruning is especially effective despite its simplicity.

\noindent\textbf{Training Objectives}\quad As another common approach to model compression, knowledge distillation offers highly effective training objectives \cite{jiao-etal-2020-tinybert}. Pruning with distillation objective shows improved performance \cite{DBLP:conf/nips/Sanh0R20,xia-etal-2022-structured}. However, in gradient-based pruning, the distillation objectives may disturb the estimation of importance scores. We propose a gradient separation strategy that uses different gradients for model optimization and importance score estimation. We show that this method leads to better performance.

 GRAIN performs task-specific pruning without additional pre-training or data augmentation.
In the experiments, we compare GRAIN with strong pruning and distillation baselines on GLUE, SQuAD, and CoNLL 2003. GRAIN notably outperforms the comparable methods in the high-sparsity regime. A demonstration of the results on MNLI is shown in Figure \ref{figure:mnli}. While keeping 5\% parameters in transformers, GRAIN maintains $93\%\sim 99\%$ performance of BERT$_{\textrm{base}}$ and $6\sim 7\times$ speedups across different tasks.
 Furthermore, GRAIN still achieves competitive results even under extreme compression where only $3\%$ transformer weights remain.

\section{Related Work}
A growing number of works have been devoted to the compression and acceleration of PLMs. Most of the works have combined multiple techniques.

\textbf {Knowledge Distillation} \cite{DBLP:journals/corr/HintonVD15} is a training technique that trains a student model to mimic the outputs and intermediate representations of the teacher model \cite{sun-etal-2019-patient}. DistilBERT \cite {Sanh2019DistilBERTAD} and TinyBERT \cite{jiao-etal-2020-tinybert} are both small BERT-like models distilled with general and task-specific distillation. MobileBERT \cite{DBLP:conf/acl/SunYSLYZ20} and KroneckerBERT \cite{tahaei-etal-2022-kroneckerbert} have designed novel structures for student models. \citet{etd} proposes to extract a subnetwork from the teacher and then perform distillation. AutoTinyBERT \cite{yin-etal-2021-autotinybert} combine distillation with neural architecture search to find optimal hyperparameters. DynaBERT \cite{DBLP:conf/nips/HouHSJCL20} apply task-specific distillation and can flexibly adjust the model size.
In this work, we only apply task-specific distillation, which consumes fewer resources.

\textbf{Structured Pruning} on PLMs remove different types of units from the models, like attention heads \cite{DBLP:conf/nips/MichelLN19}, FFN hidden dimensions \cite{liang-etal-2021-super}, blocks of weights \cite{lagunas-etal-2021-block}, MHA layers or FFN layers \cite{xia-etal-2022-structured}. Many works combine pruning with other methods. \citet{wang-etal-2020-structured} presents a structured pruning approach with low-rank factorization of weight matrices. \citet{DBLP:journals/corr/abs-1910-06360} and \citet{xia-etal-2022-structured} apply pruning with knowledge distillation. In this work, we apply matrix factorization on the embeddings and use distillation and pruning to reduce the size of transformers.

\textbf {Unstructured Pruning} removes each weight individually based on its magnitude \cite{DBLP:journals/corr/HanPTD15, DBLP:conf/iclr/ZhuG18,gordon-etal-2020-compressing}, or the score computed by first-order \cite{DBLP:conf/nips/Sanh0R20, DBLP:journals/corr/abs-1712-01312} or second-order \cite{DBLP:journals/corr/abs-2203-07259} method. Unstructured pruning yields higher sparsity models but is hard to speed up without specialized devices for sparse matrix operations. In this work, we only consider structured pruning.

Besides model compression, another group of acceleration methods is dynamic inference, where the computation cost is determined at test time \cite{DBLP:conf/iclr/FanGJ20, liu-etal-2020-fastbert, xin-etal-2020-deebert}. \citet{liu-etal-2021-ebert} and \citet{cost-eff} have proposed to integrate model compression with dynamic inference. We do not consider dynamic inference in this work and leave it for future work.

\section{Preliminaries}
\subsection{Transformers}

A Transformer block \cite{DBLP:conf/nips/VaswaniSPUJGKP17} is mainly composed of a multi-head attention (MHA) layer and a feed-forward network (FFN) layer. 

Let $\bm{X}\in \mathbb{R}^{n\times d}$ be the input sequence, where $n$ is the length, and $d$ is the hidden size. An attention head is parameterized by the matrices $\bm{W}^Q_i,\bm{W}^K_i,\bm{W}^V_i, \bm{W}^O_i \in \mathbb{R}^{d_h\times d}$. Its output is\footnote{ We omit bias terms throughout for simple presentation.}
{
\setlength{\belowdisplayskip}{3pt}
\begin{equation}\label{eq:att}
\mathrm{Att}_{i}(\bm{X}) = \mathrm{softmax}\left({\bm{Q}_i\bm{K}_i^\mathsf{T}}/{\sqrt{d}}\right)\bm{V}_i\bm{W}^O_i,
\end{equation}
}
{\small
\setlength{\abovedisplayskip}{0pt}
\begin{equation}
\bm{Q}_i=\bm{X}(\bm{W}^Q_i)^\mathsf{T}, \bm{K}_i=\bm{X}(\bm{W}^K_i)^\mathsf{T}, \bm{V}_i=\bm{X}(\bm{W}^V_i)^\mathsf{T} \nonumber,
\end{equation}
}
where $d_h$ is head size, and $i$ is the head index. An MHA layer contains $N_h=d/d_h$ attention heads
{
\begin{equation}\label{eq:mha}
\mathrm{MHA}(\bm{X}) = \sum\nolimits_i^{N_h}  \mathrm{Att}_{i}(\bm{X}) .
\end{equation}
}
Following the MHA layer is the feed-forward network layer. It consists of two linear layers and a GeLU  activation \cite{hendrycks2016gelu}
\begin{equation}
	 \textrm{FFN}(\bm{X}) = \textrm{GeLU}(\bm{X}\cdot \bm{W}_1)\cdot\bm{W_2},
	 \label{eq:ffn}
\end{equation}
where $\bm{W}_1 \in \mathbb{R}^{d\times d_{f}}$, $\bm{W}_2 \in \mathbb{R}^{d_{f}\times d}$, and $d_{f}$ is the intermediate hidden size. Typically  $d_{f} > d$.

A transformer block contains other components, such as LayerNorm and residual connection, but they only take up a few parameters.

\begin{figure*}[th]
  \includegraphics[width=\linewidth]{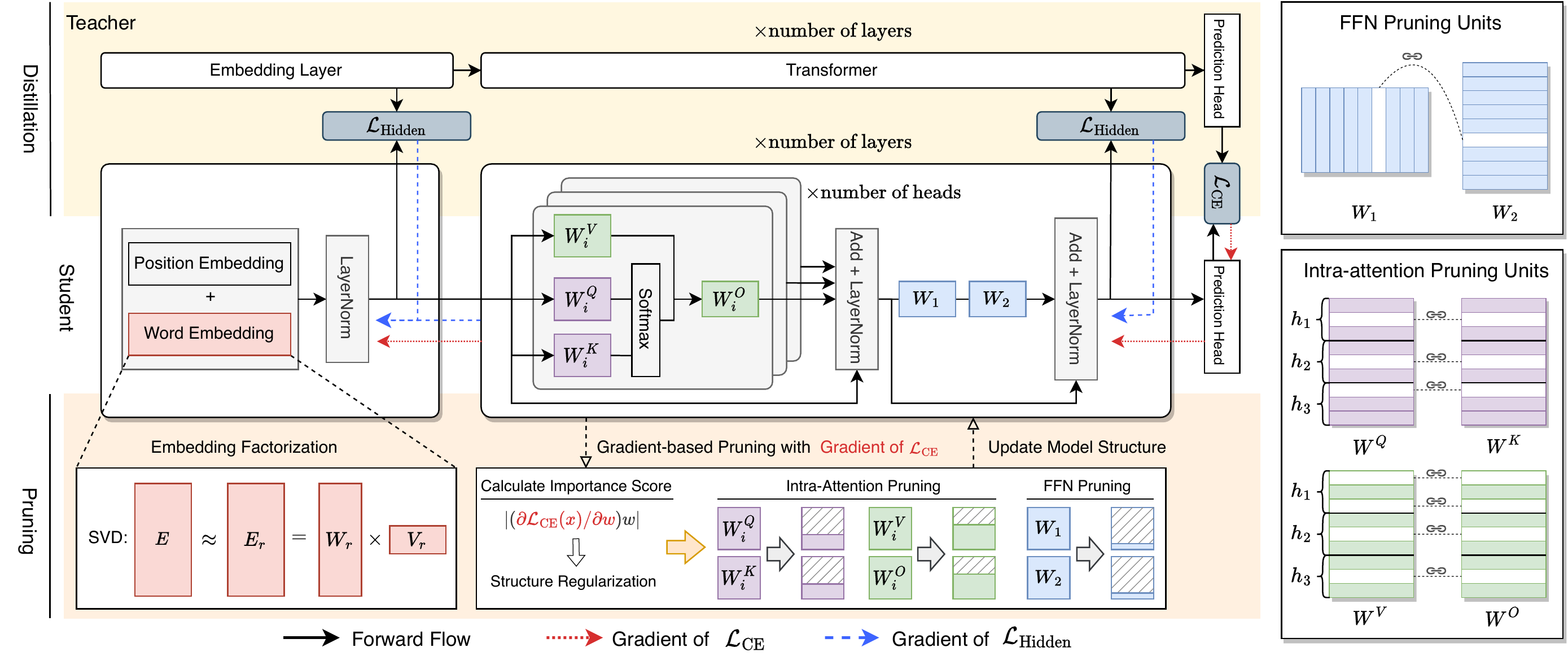} 
  \caption{An overview of GRAIN. Left: The distillation and pruning methodology of the student model. Right: The pruning units in intra-attention pruning. We show three heads for illustration. Each head can be pruned to a different size. The pruned units are in white. The dashed lines indicate that the connected rows or columns should be removed together.}
  \label{figure:GRAIN}
\end{figure*}

\subsection{Gradient-based Pruning}

Gradient-based pruning \cite{DBLP:conf/nips/MichelLN19} defines the importance score of a pruning unit $w$ as the variation of the loss with respect to the unit:
\begin{equation}\label{ISeq}
  \mathrm{IS}(w) = \mathbb{E}_{x\sim X}\left\lvert \frac{\partial \mathcal{L}(x)}{\partial w}w \right\rvert,
\end{equation}
where $X$ is the data distribution. The term in the absolute value is the first-order Taylor approximation of the loss $\mathcal{L}$ around $w=0$. To apply \eqref{ISeq} in PLM pruning, $w$ should be set accordingly. 
For example, by setting $w$ to $\bm{W}^O_i$, Equation \eqref{ISeq} gives the importance score of the head $h_i$; by setting $w$ to the $i$-th row of $\bm{W}_2$, Equation \eqref{ISeq} gives the importance score of the $i$-th FFN hidden dimension.
A lower importance score implies that the loss is less sensitive to the unit. The pruning units are sorted and then pruned in the order of increasing scores.

\section{Methodology}
GRAIN performs task-specific intra-attention pruning together with knowledge distillation. The overview of GRAIN is depicted in Figure \ref{figure:GRAIN}.
Following previous works, we only include the encoder in counting the model size unless otherwise specified. 
We refer to the size of the pruned model relative to the unpruned model as \textit{model density}:
$$
\textrm{model density} = \frac{\texttt{SizeOf(} \textrm{pruned model}\texttt{)}}{\texttt{SizeOf(} \textrm{original model}\texttt{)}}.
$$
 \textit{Sparsity} is equal to one minus model density.

\subsection{Intra-attention Pruning}
\subsubsection{Intra-attention Pruning Units}
FFN hidden dimensions and attention heads are common pruning units in  PLM pruning studies. These pruning units have been treated as atomic in structured pruning. However, attention heads include finer pruning units and are not really atomic. Equation \eqref{eq:mha} shows that the output of an MHA layer is the sum of individual heads, so different heads can be pruned independently. To be specific, We can remove the rows of the matrices $\bm{W}^Q_i,\bm{W}^K_i,\bm{W}^V_i, \bm{W}^O_i$ to reduce head size. Further, from Equation \eqref{eq:att}, we see that the output dimensions of $\bm{W}^Q_i,\bm{W}^K_i$ and the input dimensions of $\bm{W}^V_i, \bm{W}^O_i$ can be different. It gives another freedom to set the dimensions of attention heads.

Based on the above observation, we introduce two kinds of intra-attention pruning units: query units, namely the rows of $\bm{W}^Q_i,\bm{W}^K_i$; and value units, namely the rows of $\bm{W}^V_i, \bm{W}^O_i$. We keep FFN hidden dimensions but discard attention heads as the pruning units since the intra-attention pruning units are more structurally fundamental. Each pruning unit takes $2d$ parameters. The new set of pruning units greatly expands the structure space.

In the actual implementation \cite{wolf-etal-2020-transformers}, the parameters of all heads in an MHA layer are gathered and stored in four large matrices $\bm{W}^Q,\bm{W}^K,\bm{W}^V, \bm{W}^O \in \mathbb{R}^{d \times d}$. The parameters of the $i$-th head are stored in the rows $(i,i+d_h)$. We prune query and value units from large matrices by removing corresponding rows. The pruning units are illustrated in the right part of Figure \ref{figure:GRAIN}.

\subsubsection{Structure Regularization}
\label{section:SR}

Since intra-attention pruning removes the units inside attention heads, it tends to generate models with many small heads of different sizes, but the total number of heads can still be large. We refer to this kind of structure as fragmented (see the upper panel in Figure \ref{figure:structure} for an example). The fragmented structure has low efficiency on devices like GPUs since there are still many attention modules left in the model, and these heads are hard to parallelize.

To remedy this, we introduce \textbf{Structure Regularization} (\textbf{StructReg} for short) to encourage generating less fragmented structures. Intuitively, to avoid small heads, the pruning process should first prune the units in the small heads and make them empty, which can then be safely removed. 

To be general, we define $D(\mathcal{M},\mathcal{W})$ as the density of a set of pruning units $\mathcal{W}$  in module $\mathcal{M}$, i.e., the ratio of the remaining units in $\mathcal{M}$. The regularized importance score of a unit $w\in\mathcal{W}$ is:
\begin{equation}\label{eq:sr}
	\textrm{IS}^r(w) = \textrm{IS}(w)\cdot \tanh(D(\mathcal{M},\mathcal{W})/\alpha) ,
\end{equation}
where $\alpha$ is the regularization strength. The lower the density of the units in $\mathcal{M}$, the lower the regularized scores of the units. Hence, the units in low-density modules will be pruned with priority until all the units in $\mathcal{M}$ have been pruned, leaving fewer low-density modules in the pruned model.

StructReg can be applied on different levels by choosing different $\mathcal{M}$s and $\mathcal{W}$s.
We apply it to intra-attention structures. We set $\mathcal{M}$ to each attention head and $\mathcal{W}$ to the value units in $\mathcal{M}$. Heads with fewer value units will be pruned with priority until empty, resulting in fewer small heads.

\subsection{Knowledge Distillation}
\textbf{Distillation Objectives}\quad Knowledge distillation provides effective objectives for transferring knowledge from a large model to a small model. The most simple distillation objective involves a cross-entropy loss between the student's and the teacher's prediction probabilities
 \begin{align}
 \mathcal{L}_{\textrm{CE}} = \bm{p}^{(T)}_{\tau}\cdot \log \bm{p}^{(S)}_{\tau},
 \label{eq:ce}
 \end{align}
where $T$ and $S$ denote \textit{teacher} and \textit{student} respectively, and $\bm{p}_{\tau}=\textrm{softmax}(\bm{z}/\tau)$ is the scaled probability with temperature $\tau$ and logits $\bm{z}$. By integrating logits distillation with hidden layer representation distillation \cite{jiao-etal-2020-tinybert,DBLP:conf/acl/SunYSLYZ20}, the performance of knowledge distillation can be further improved:

\begin{align}
	\mathcal{L}_{\textrm{Hidden}} =  \sum_{(i,j)\in \mathcal{I}} \textrm{MSE}(\bm{H}^{(S)}_i \bm{W}_i, \bm{H}_j^{(T)}),
	\label{eq:hidden}
\end{align}
where $\mathcal{I}$ is the set of layer index pairs, $ \bm{H}_{i} (i>0)$ is the hidden states from the $i$-th transformer block ($\bm{H}_0$ is the output from the embedding layer), and $\bm{W}_i$ is a trainable linear mapping. We employ the sum of $\mathcal{L}_{\textrm{CE}}$ and $\mathcal{L}_{\textrm{Hidden}}$ as the total loss.

\noindent\textbf{Gradient Separation}\quad When applying distillation with gradient-based pruning, the hidden layer matching loss $\mathcal{L}_{\textrm{Hidden}}$ should be treated carefully. 
In gradient-based pruning, the units are pruned based on how significantly they affect the model predictions. Thus, the importance score should be calculated solely from the cross-entropy loss, and we should avoid the gradients from other losses like $\mathcal{L}_{\textrm{Hidden}}$ affecting the estimation of the importance scores. Therefore, we propose to use the gradient from $\mathcal{L}_{\textrm{CE}}$ for model optimization and importance score computation, while using the gradient from  $\mathcal{L}_{\textrm{Hidden}}$  only for model optimization. We call this strategy \textbf{gradient separation} (GS). The gradient flows of different losses are illustrated in Figure \ref{figure:GRAIN}.

\subsection{Iterative Gradient-based Pruning}
\label{sec:procedure}

\textbf{Iterative Pruning}\quad Similar to \citet{DBLP:conf/nips/Sanh0R20}, we take an iterative approach to prune the model, i.e., the model size is gradually reduced during fine-tuning. We denote the total training steps as $N$ and the current step as $i$. The model is pruned to the density $s(t)$ at every step, where $s(t)$ is the density scheduler as a function of the training percentage $t = i/N \in [0,1]$. We will give the exact form of $s(t$) shortly. Notice that in the standard gradient-based \pageref{•}runing, the importance score is estimated from all the examples in the dataset $X$ (see Equation \eqref{ISeq}). It would be impractical to estimate the score at every step. Therefore we define an exponentially smoothed importance score $\overline{\textrm{IS}_i}(w)$ which can be computed efficiently during training and used for pruning at step $i$:
\begin{equation}\label{smooth}
	\overline{\textrm{IS}_i}(w) = \beta\cdot\overline{\textrm{IS}_{i-1}}(w) + (1-\beta)\cdot\textrm{IS}_i(w),
\end{equation}
where $\textrm{IS}_i(w)$ is the importance score of the pruning unit $w$ calculated  with a single batch at step $i$, and $\beta$ is the smoothing factor.
The smoothed score avoids the large variance and leads to more stability. Equation \eqref{smooth} can also be applied on the regularized score simply by  replacing $\textrm{IS}(w)$ with $\textrm{IS}^r(w)$.

\begin{figure}[t]
  \includegraphics[width=\linewidth]{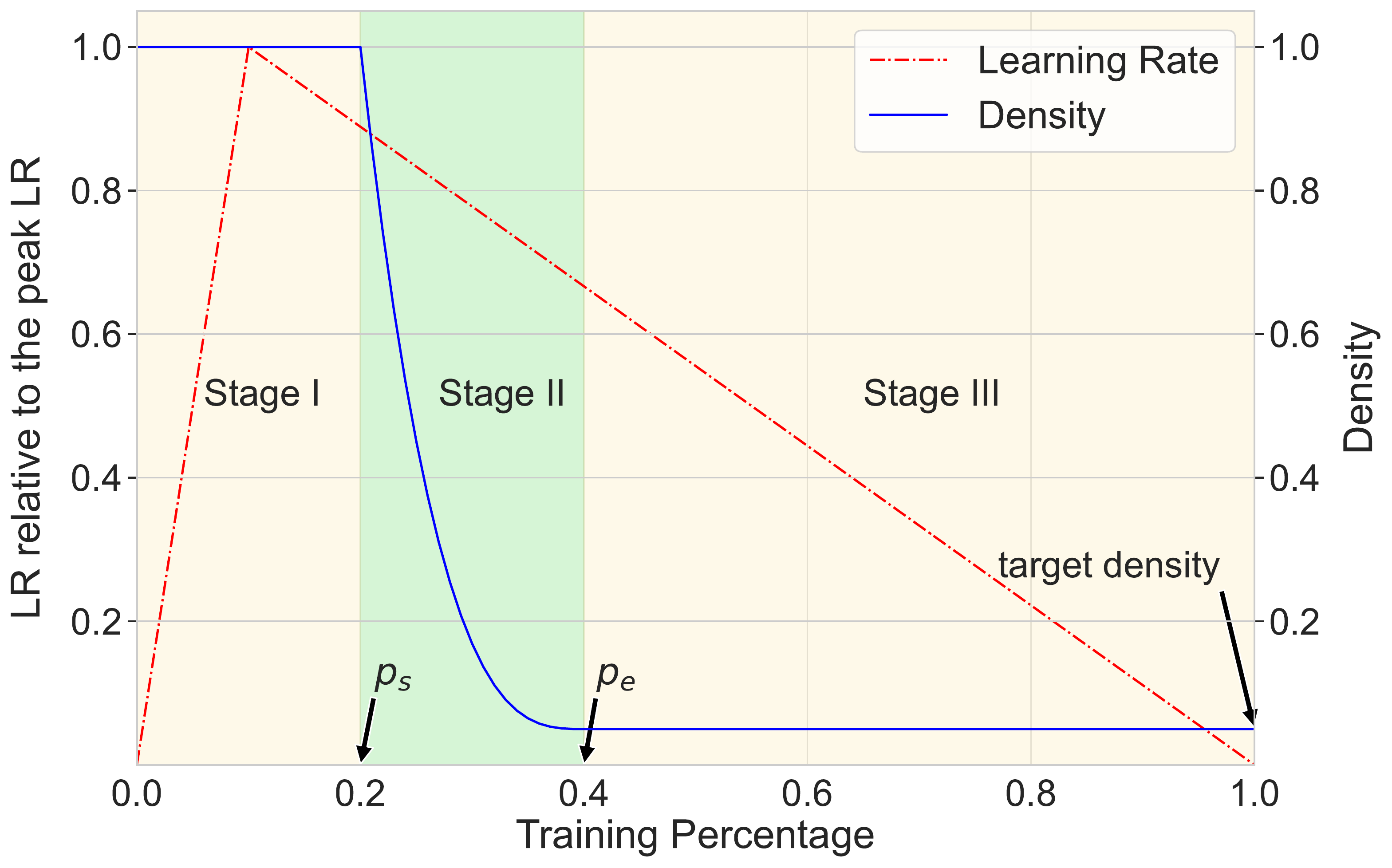} 
  \caption{The training stages of GRAIN.}
  \label{figure:procedure}
\end{figure}

\noindent\textbf{Scheduling}\quad Following \citet{DBLP:conf/iclr/ZhuG18}, we use a cubic density scheduler $s(t)$
{
\begin{equation}
\label{eq:cubic}
\begin{cases}
1 & 0\le t<p_s \\
 s_f + (1-s_f)(1-\frac{t-p_s}{p_e-p_s})^3 & p_s\le t\le p_e\\
 s_f & p_e < t \le 1 \nonumber \\
\end{cases}.
\end{equation}
}
The complete process can be divided into three stages, as depicted in Figure \ref{figure:procedure}. The first stage is the warm-up stage. We train the student model for $N p_s$ steps with the distillation objective, where $0<p_s<1$ is a hyperparameter.   
In the second stage, we gradually prune the model with distillation for $ N (p_e - p_s)$ steps. The model density $s$ decreases from the initial density ($100\%$) to the target density $s_f$ following the schedule. In the last stage, the model structure is fixed,  and we continually train the model with distillation to recover performance \cite{DBLP:conf/nips/Sanh0R20, DBLP:conf/iclr/ZhuG18}. The three stages take place consecutively, and the whole process is done in a single run of fine-tuning.

\subsection{Embedding Factorization}
The pruning mentioned above reduces the parameters in the transformers, while another large fraction of the parameters stored in the word embedding matrix is untouched. 
We apply singular value decomposition (SVD) to reduce the embedding size. SVD decomposes the word embedding matrix $\bm{E}\in\mathbb{R}^{q\times d}$ as
$
\bm{E} = \bm{U}\bm{\Sigma}\bm{V}
$,
where $q$ is the vocabulary size and $d$ is the hidden size,
$\bm{U}\in \mathbb{R}^{q \times d}$, $\bm{V}\in \mathbb{R}^{d \times d}$ and $\bm{\Sigma}$ is a diagonal matrix composed of singular values. $\bm{E}$ can be approximated as $\bm{E}_r$ by selecting top $r$ singular values and corresponding $r$ rows from $\bm{U}$ and $\bm{V}$ 
\begin{equation}
\bm{E}\approx\bm{E}_r = \bm{U}_r \Sigma_r	 \bm{V}_r = \bm{W}_r \bm{V}_r,
\end{equation}
where $\bm{W}_r\in\mathbb{R}^{q \times r}$ and $\bm{U}_r\in \mathbb{R}^{r \times d}$. The original embedding $\bm{E}$ is now replaced by $\bm{W}_r$ and $\bm{V}_r$. The embedding size is reduced from $qd$ to $(q+d)r$.

Embedding factorization has little effect on latencies but significantly reduces model sizes. 
Some works \cite{xia-etal-2022-structured,lagunas-etal-2021-block} do not prune embeddings. We also conduct experiments without embedding factorization for comparison. We name this setting as \textbf{GRAIN w/o EF}.

\begin{figure*}[th]
  \includegraphics[width=\linewidth]{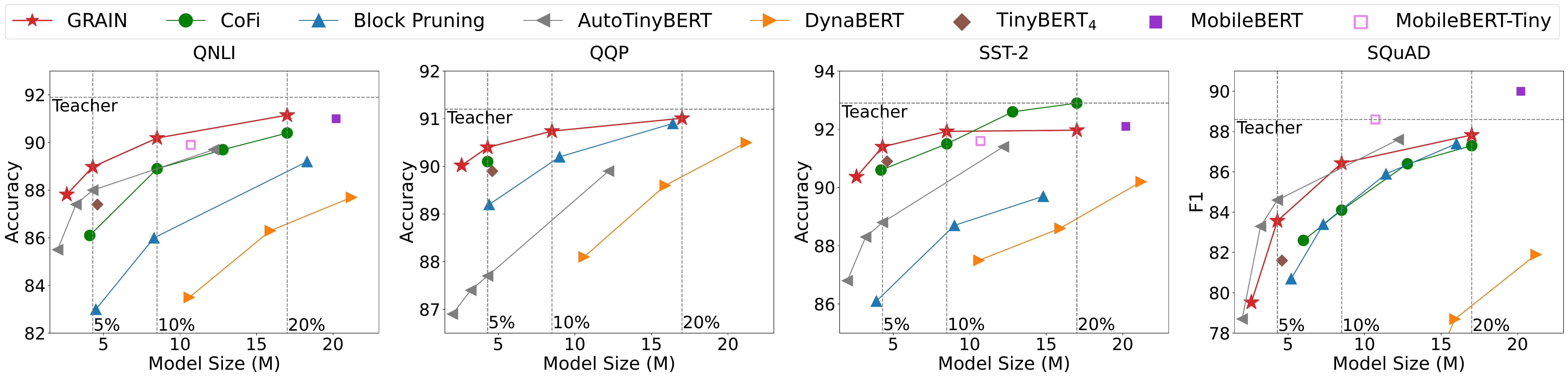} 
  \caption{Performance of GRAIN and baseline methods at different model densities. The results of CoFi, AutoTinyBERT, and MobileBERT are taken from the original papers and the CoFi repository. Since the precise sizes of CoFi models around density 10\%, 15\%, and 20\% are unknown, we use 10\%, 15\%, and 20\% as estimates instead. }
  \label{figure:main_results}
\end{figure*}

\begin{table*}[ht]
\centering
\small
\begin{tabular}{lcccccccc}
\toprule
\textbf{Model} &
  \begin{tabular}[c]{@{}c@{}}\textbf{QNLI}\\ {(Acc)}\end{tabular} &
  \begin{tabular}[c]{@{}c@{}}\textbf{MNLI}\\ {(m/mm Acc)}\end{tabular} &
  \begin{tabular}[c]{@{}c@{}}\textbf{QQP}\\ {(Acc)}\end{tabular} &
  \begin{tabular}[c]{@{}c@{}}\textbf{SST-2}\\ {(Acc)}\end{tabular} &
  \begin{tabular}[c]{@{}c@{}}\textbf{SQuAD}\\ {(F1 / EM)}\end{tabular} &
  \begin{tabular}[c]{@{}c@{}}\textbf{CoNLL-03}\\ {(F1)}\end{tabular} &
  \begin{tabular}[c]{@{}c@{}}\textbf{Model}\\ {\textbf{Size}}\end{tabular} &
  \begin{tabular}[c]{@{}c@{}}\textbf{Total}\\ {\textbf{Size}}\end{tabular} \\ \midrule
BERT$_{\textrm{base}}$ (teacher) & 91.9 & 84.7 / 85.0 & 91.2 & {92.9} & 88.6 / 81.1 & 91.2 & 85.1M        & 108.9M \\ \midrule
\rowcolor[HTML]{E4E4E4}
\textit{5\% Model Density}        &      &           &           &      &           &       &              &        \\
TinyBERT$_4$$^\dagger$           & 87.4 & 80.9 / 81.9 & 89.9 & 90.9 & 81.6 / 71.9 & 84.9  & 4.7M (5.5\%) & 14.6M \\
AutoTinyBERT$^{\mathsection}$           & 88.0 & 79.4 / \;\;-\;\;\;  & 87.7 & 88.8 & \textbf{84.6} / \;\;-\;\;\;& -  & 4.3M (5.0\%) & 14.5M \\

Block Pruning$^\dagger$           & 83.0 & 78.9 / 78.6 & 89.2 & 86.1 & 80.7  / 71.0 & 84.0  & 4.6M (5.4\%) & 28.8M \\

CoFi (reimpl.)$^\dagger$                   & 85.3 &  79.8 / 79.6     & 89.8     & 89.8 &  79.0 / 69.2     & 85.0  & 4.2M (4.9\%) & 28.2M   \\
CoFi$^{\mathsection}$                   & 86.1 &  80.6 / 80.7     & 90.1     & 90.6 &  82.6 / \;\;-\;\;\;     & -  & 4.7M (5.5\%)$^\ddagger$ & 29.0M$^\ddagger$   \\  \arrayrulecolor{black!30}\midrule
GRAIN                       & 89.0 & 82.2 / \textbf{82.5} & {90.4} & 91.4 & {83.6 / 73.7} & \textbf{88.3} & 4.3M (5.0\%) & 10.7M  \\ 
GRAIN w/o EF                & \textbf{89.1} & \textbf{82.4} / 82.2 & \textbf{90.5} & \textbf{91.6} & 83.4 / 73.2 & \textbf{88.3} & 4.3M (5.0\%) & 28.1M  \\ \arrayrulecolor{black!100}\midrule
\rowcolor[HTML]{EAEAEA}
\textit{3\% Model Density}                 &      &           &           &      &           &       &              &        \\
GRAIN                       & \textbf{87.8} & 80.7 / 81.1 & 90.0 & 90.4 & \textbf{79.5 / 68.4} & 86.8 & 2.6M (3.0\%) & 9.0M   \\
GRAIN w/o EF                & 87.6 & \textbf{81.0} / \textbf{81.2} & \textbf{90.2} & \textbf{91.0} & 79.0 / 67.3 & \textbf{87.2} & 2.6M (3.0\%) & 26.4M  \\ \bottomrule
\end{tabular}
\caption{Results of GRAIN and the baselines at model density around 5\% and 3\%. The best results are in bold. We average the sizes across different tasks in counting the model and total size. $^\mathsection$: The results are taken from the original papers.
$\dagger$: The results are obtained by our reimplementation with the released code.
$\ddagger$: The size is averaged excluding QQP since there are no public CoFi model checkpoints for the QQP task.
}
\label{table:main_results}
\end{table*}

\section{Experiments}
\label{section:Experiments}
\subsection{Experiment Setup}
\noindent\textbf{Datasets}\quad We evaluate our approach on machine reading comprehension SQuAD 1.1 \cite{rajpurkar-etal-2016-squad}, named entity recognition CoNLL 2003 \cite{conll2003}, and four classification tasks (SST-2, QNLI, MNLI, and QQP) that have relative large training data from GLUE benchmark \cite{gluebenchmark}. Details are summarized in Appendix \ref{sec:dataset}. We report the results on the development sets of GLUE and SQuAD and the results on the test set of CoNLL 2003.

\noindent\textbf{Training Settings}\quad
We use BERT$_{\textrm{base}}$ as the backbone model.\footnote{We also experiment with RoBERTa \cite{liu2019roberta} and Chinese-RoBERTa-wwm-ext \cite{cui-etal-2021-pretrain} on Chinese tasks. See Appendix \ref{app:results} for details.} We first fine-tune the teachers for each task, then train and prune the students following the procedure in Section \ref{sec:procedure}. The target model densities range from $3\%$ to $20\%$. We list the model size  and the total size (with embeddings and classifiers) for reference. We report the mean score of 3 runs with different random seeds. See Appendix \ref{sec:hyperparameters} for training details and costs.

\noindent\textbf{Baselines}\quad
We compare our proposed method with \textbf{CoFi} \cite{xia-etal-2022-structured}, \textbf{Block Pruning} \cite{lagunas-etal-2021-block}, \textbf{TinyBERT}$_4$ \cite{jiao-etal-2020-tinybert} and \textbf{DynaBERT} \cite{DBLP:conf/nips/HouHSJCL20}. We also list the results of \textbf{AutoTinyBERT} \cite{yin-etal-2021-autotinybert} and \textbf{MobileBERT} \cite{DBLP:conf/acl/SunYSLYZ20}. However, they are not directly comparable to {GRAIN} since they have been distilled from different teacher models and pre-trained extensively, consuming much more computation. Following \citet{xia-etal-2022-structured}, we re-implement TinyBERT$_4$ and DynaBERT without task-specific data augmentation for a fair comparison. We also re-implement CoFi and Block Pruning with their public code, and choose \textit{Hybrid Filled} approach as the Block Pruning baseline. We use the same teachers in training for GRAIN, TinyBERT$_4$, CoFi, and Block Pruning.

\subsection{Main Results}

\begin{table}[t]
\centering
\resizebox{\linewidth}{!}{

\begin{tabular}{@{}lccc@{}}
\toprule
\textbf{Method}           & \textbf{QNLI} & \textbf{SST-2} & \textbf{SQuAD} \\ \midrule
GRAIN                     & 89.0          & 91.4           & 83.6           \\
GRAIN w/o EF              & 89.1          & 91.6           & 83.4           \\
$-$ StructReg             & 89.4          & 92.2           & 83.1           \\
\ \ \ $-$ GradSep                 & 89.3          & 92.0           & 82.8           \\
\ \ \ \ \ \ $-$ Hidden Layer Loss & 86.1          & 88.1           & 80.3           \\
\ \ \ \ \ \ $-$ Importance Scores & 82.3          & 88.0           & 65.7           \\ \bottomrule
\end{tabular}

}
\caption{Ablation results at 5\% model density.}
\label{table:ablation}
\end{table}

In Figure \ref{figure:mnli}  and Figure \ref{figure:main_results}, we show the scores of GRAIN and the baseline methods on various downstream tasks with model densities ranging from $3\%$ to $20\%$. Table \ref{table:main_results} summarizes the detailed results at densities 5\% and 3\%.\footnote{Please refer to Table \ref{table:main_results_appendix} in Appendix \ref{app:results} for detailed results of GRAIN at higher model densities.} We see that GRAIN outperforms baselines in the majority of tasks on a wide range of model sizes. GRAIN outperforms TinyBERT$_4$ and Block Pruning on all tasks and outperforms CoFi except on SST-2 at relatively high density. Especially, in the low-density regime, GRAIN exhibits notable advantages over other methods. Under extreme compression at density $3\%$, GRAIN (2.6M) can match TinyBERT (4.7M) and CoFi (4.7M) on most tasks, despite having fewer parameters. In addition, compared to MobileBERT and AutoTinyBERT, which require general pre-training and use different teachers than GRAIN's, although not directly comparable, GRAIN shows promising results with less computation.

In Table \ref{table:main_results}, we show the results of GRAIN without embedding factorization (\textbf{GRAIN w/o EF}). One can see that the pruned models do not always benefit from having large embeddings. On SQuAD, the factorized embedding leads to improved performance, while on SST-2, a large embedding matrix is better. 
However, the gaps at model density 5\% are closer than those at model density 3\%, indicating that embedding factorization has more minor impacts on larger pruned models.

We also measure the latency of GRAIN and find that GRAIN achieves competitive speedups when compared with other methods. Please refer to Appendix \ref{sec:speed_results} for more details.

To summarize the above, GRAIN is efficient and effective for compressing pre-trained language models on a wide range of downstream tasks.

\subsection{Ablation Study}

\begin{table}[t!]
\centering
\resizebox{\linewidth}{!}{
\begin{tabular}{@{}lccc@{}}\toprule
 \textbf{Units} & (\textbf{FFN}, \textbf{Heads}) & \textbf{QNLI}          & \textbf{SQuAD}              \\ \midrule
Intra+FFN           & (3.5\%, 7,9\%)       & \textbf{89.0} &  \\
Intra+FFN           & (3.5\%, 8.0\%)       &  & \textbf{83.6} \\ \midrule
Heads+FFN           & (5.0\%, 5.0\%)           & 87.3          & 77.3          \\
Heads+FFN           & (3.75\%, 7.5\%)       & 88.2          & 79.2          \\
Heads+FFN           & (3.0\%, 9.0\%)       & {\underline {88.5}}          & {\underline{81.4}}         \\
Heads+FFN           & (2.5\%, 10\%)        & {\underline {88.5}}    & 80.9    \\
Heads+FFN           & (1.5\%, 12\%)        & 88.2          & 80.8          \\ \bottomrule
\end{tabular}
}
\caption{Comparison between different pruning units at 5\% model density. Heads+FFN denotes pruning with attention heads and FFN hidden dimensions. Intra+FFN denotes pruning with intra-attention units and FFN hidden dimensions. The best results are shown in bold. The best results with Heads+FFN are underlined.}
\label{table:headspruning}
\end{table}

We apply ablations on GRAIN w/o EF to study the effect of each component, as listed in Table \ref{table:ablation}. 

Firstly, The impact of removing StructReg varies depending on the task, with performance either increasing or decreasing. We defer the detailed  discussion on StructReg to Section \ref{sec:analysis}.

Secondly, we remove gradient separation (GradSep), so the importance scores are influenced by gradients from both $\mathcal{L}_{\textrm{Hidden}}$ and  $\mathcal{L}_{\textrm{CE}}$. The performance on different tasks drops more or less, and SQuAD is most notably affected. The results indicate that the gradients from the hidden layer loss  $\mathcal{L}_{\textrm{Hidden}}$ have an impact on the pruning process, and it would be more beneficial to exclude it from the estimation of importance scores.

Thirdly, we remove the hidden layer loss $\mathcal{L}_{\textrm{Hidden}}$, so knowledge distillation only optimizes the cross-entropy objective $\mathcal{L}_{\textrm{CE}}$. The performance drops significantly, showing the necessity to use both objectives for obtaining effective pruned models.

Lastly, we investigate if gradient-based pruning is necessary and effective. To ablate gradient-based pruning, we generate random scores instead of gradient-based scores at each pruning step and keep all other settings unchanged, so the models are randomly pruned. The results are displayed in the last line in Table \ref{table:ablation}. The random structures resulted in inferior results, proving the superiority of the structures found by gradient-based pruning. 
Thus both pruning and distillation are crucial components.

\begin{figure*}[t!]
\centering
  \includegraphics[width=0.245\linewidth]{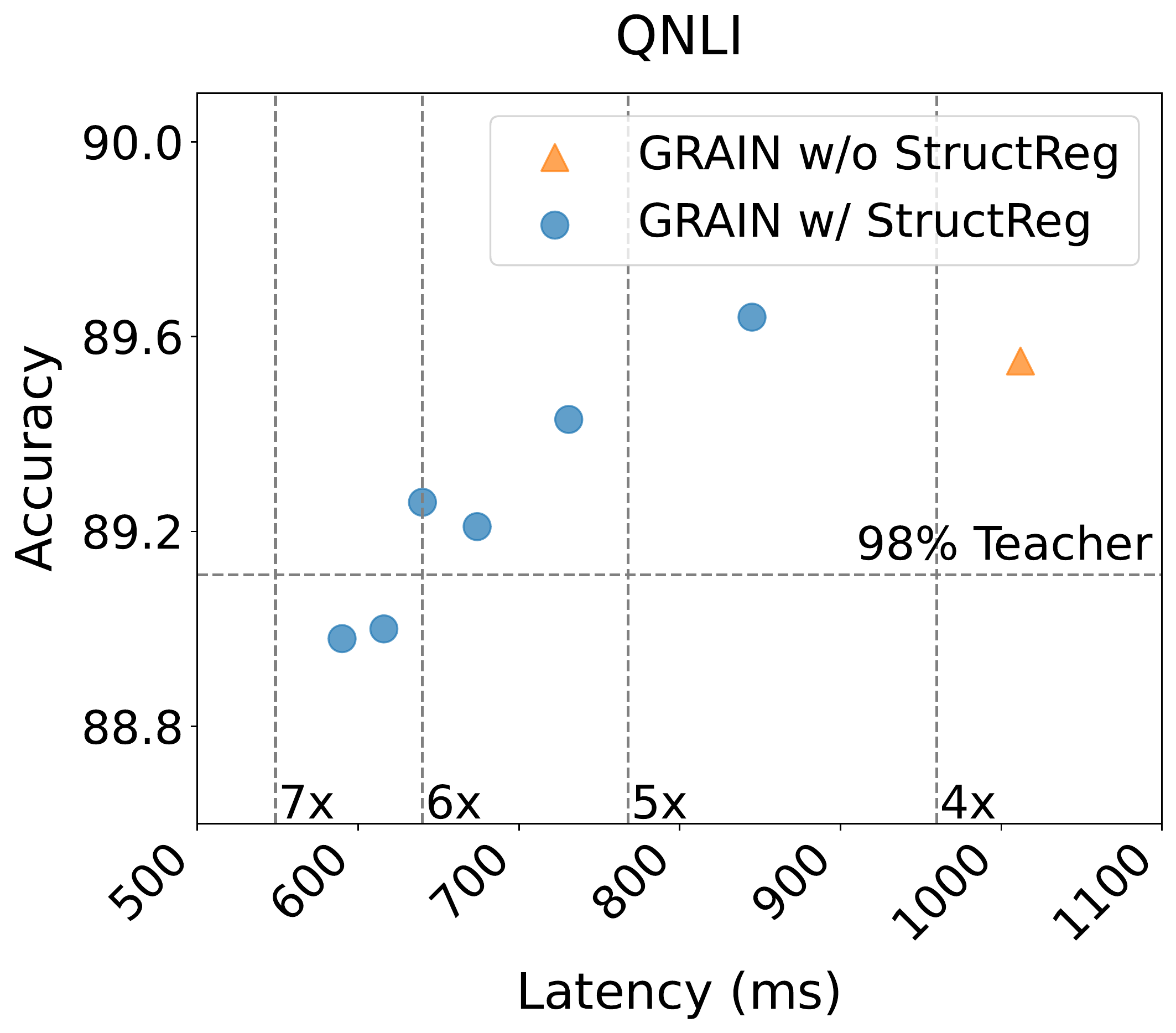}
  \includegraphics[width=0.245\linewidth]{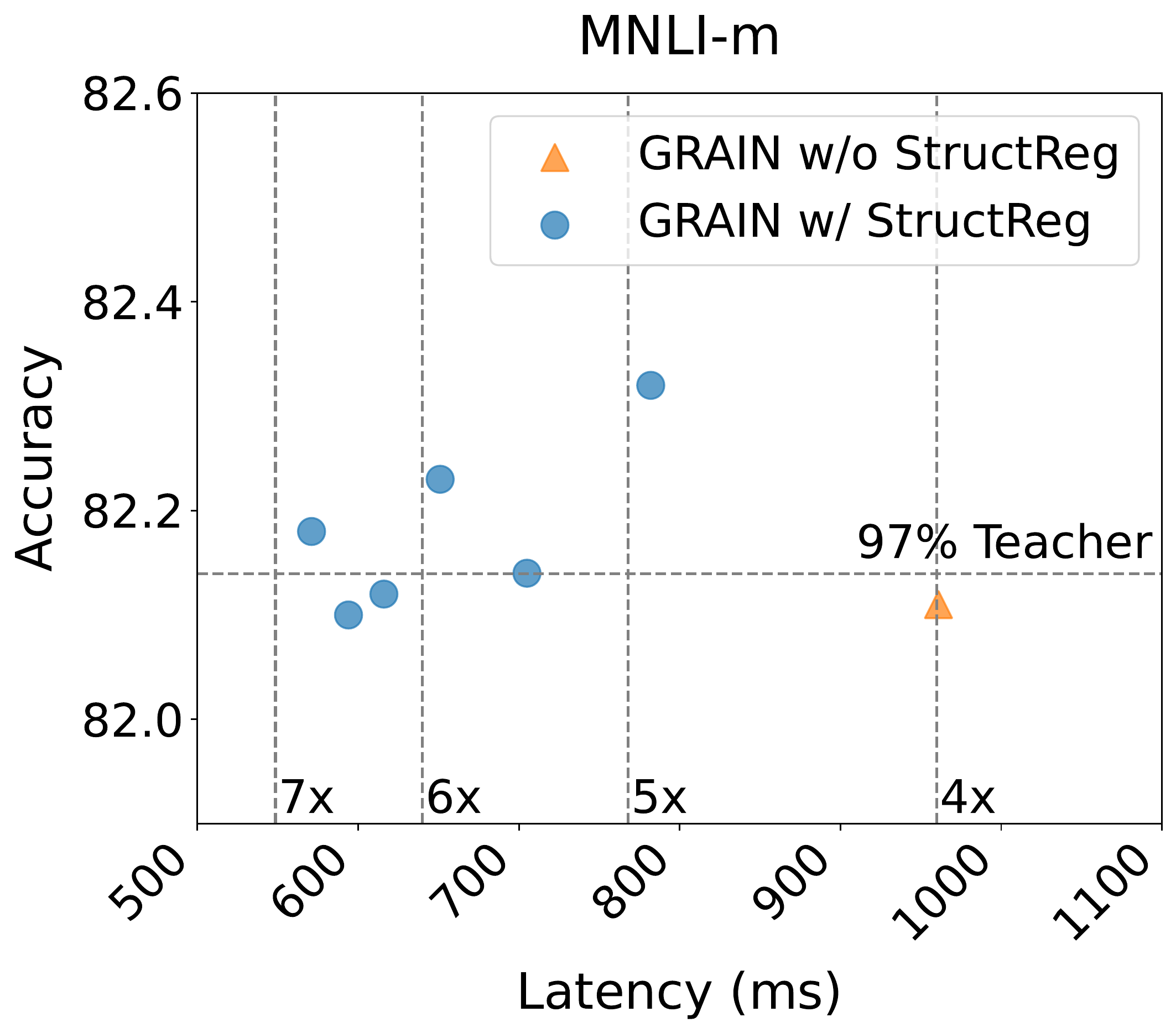} 
  \includegraphics[width=0.245\linewidth]{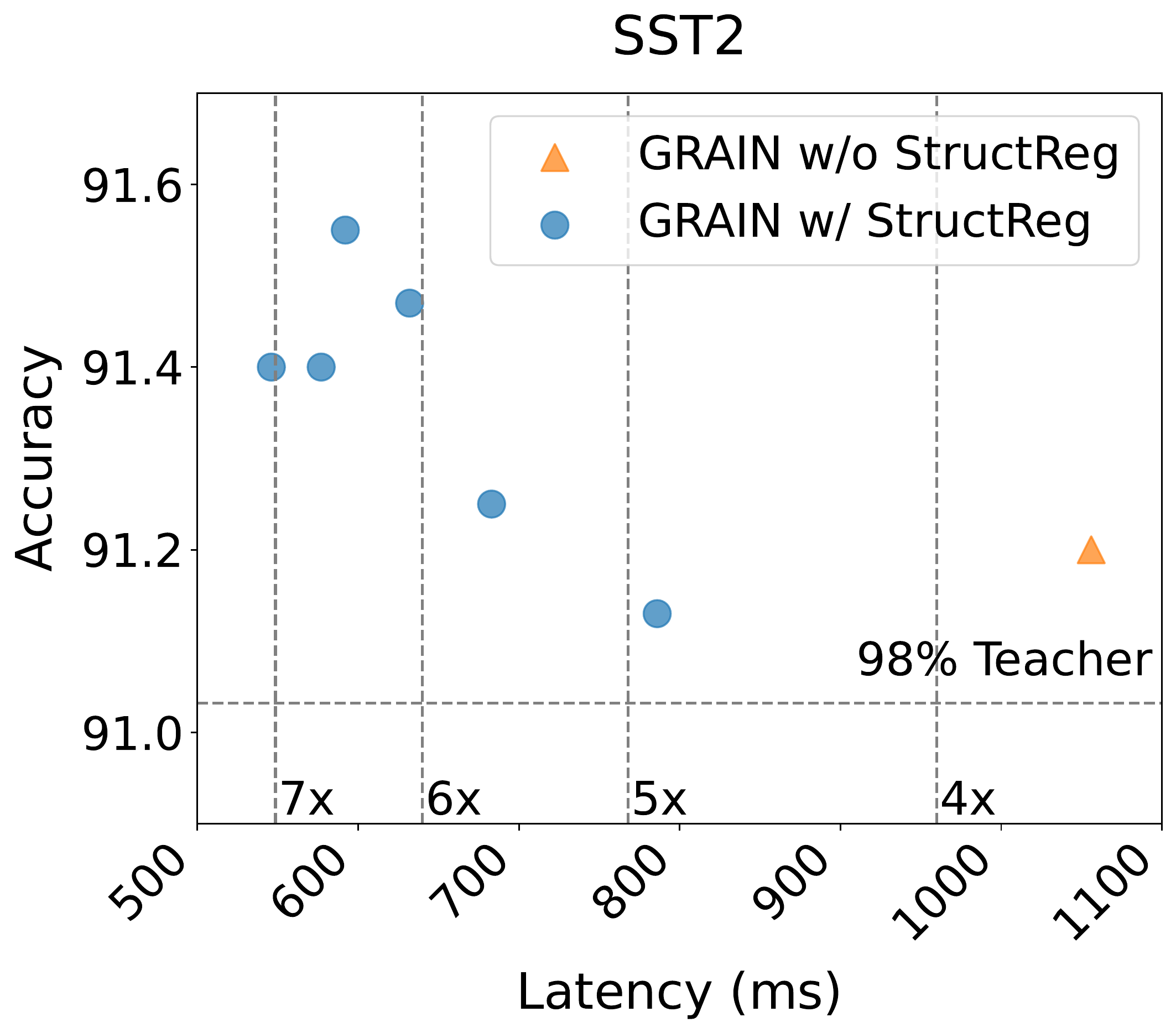}
  \includegraphics[width=0.245\linewidth]{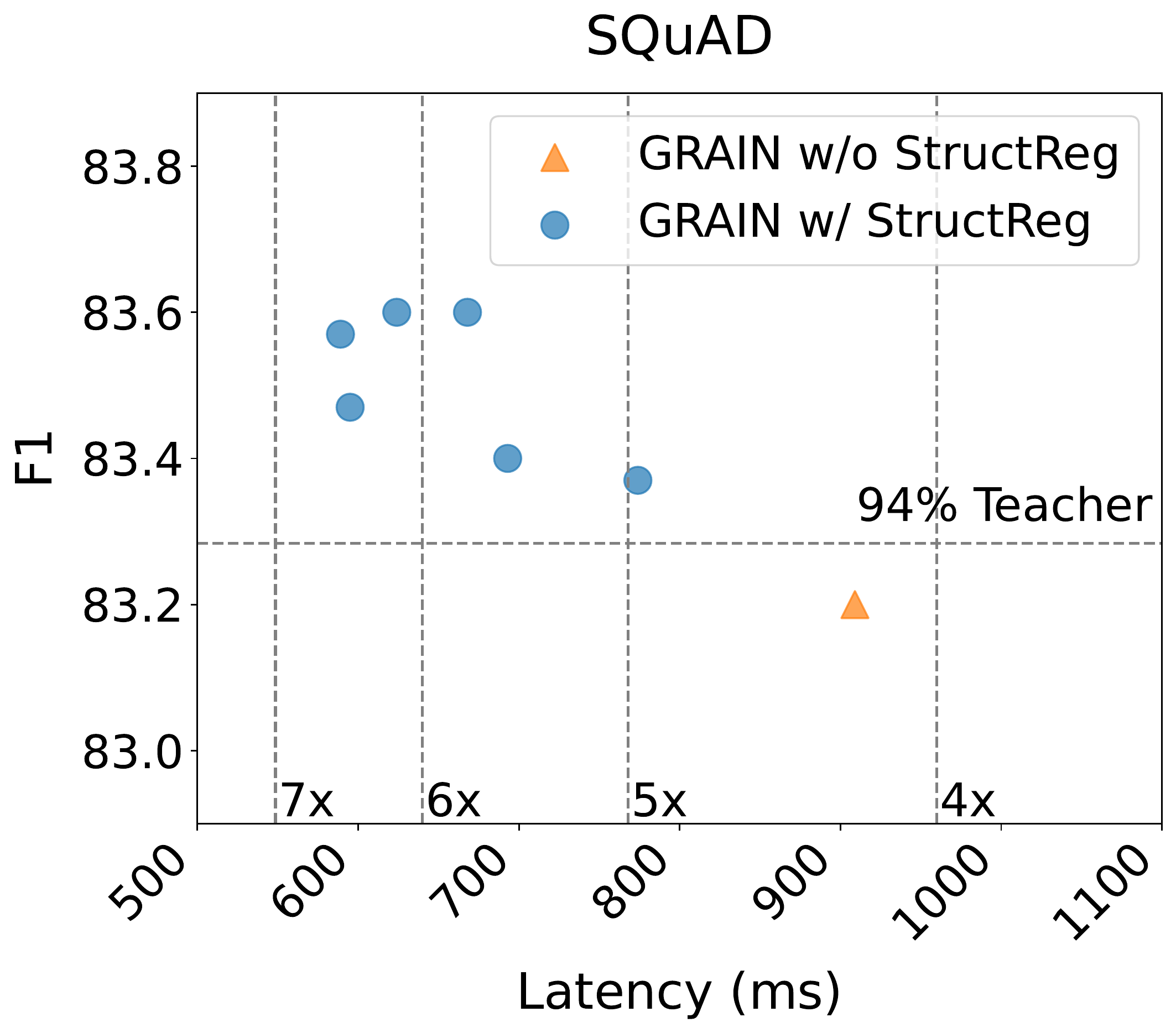}
  \caption{Performance and latencies under different structure regularization strengths at density $5\%$. The orange marker denotes $\alpha=0$, and the blue markers denote $\alpha=0.05, 0.1, 0.15, 0.2, 0.25, 0.3$ from right to left. The vertical lines indicate the speedups relative to BERT$_{\textrm{base}}$ (teacher). The horizontal line indicates the teacher's performance. }
  \label{figure:sr}
\end{figure*}

\begin{figure}[t]
\centering
  \includegraphics[width=\linewidth]{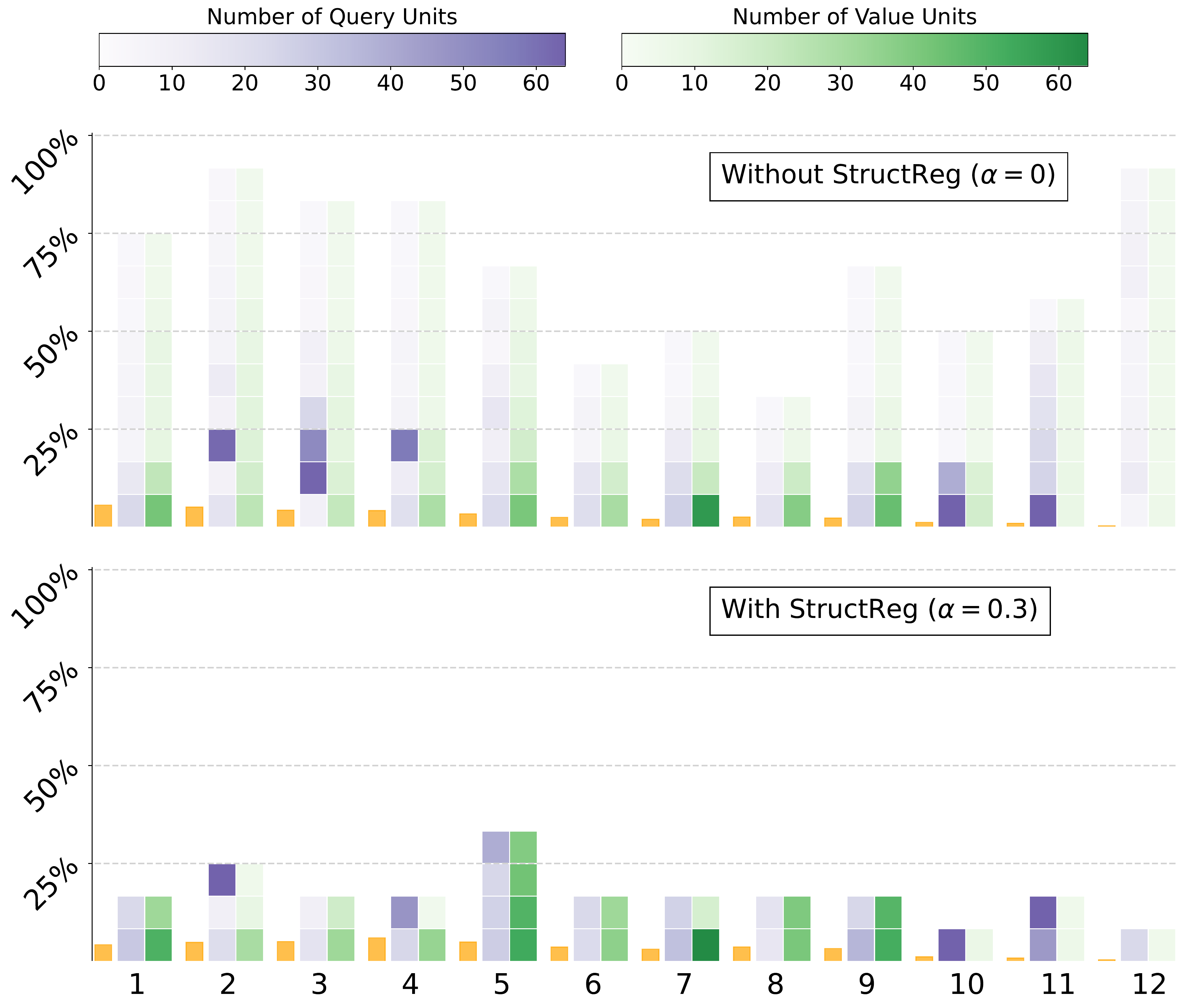} 

  \caption{The structures of the pruned models at $5\%$ density on QNLI. Orange bars denote the remaining units in FFN layers. Each purple (green) cell denotes a remaining pair of query/key (value/output) matrices in a head. The darker the cell, the more units remain.}
  \label{figure:structure}
\end{figure}

\subsection{Analysis}\label{sec:analysis}
We first compare the effects of different pruning units. Then we look into the structures of pruned models to better understand our method.

\noindent\textbf{Attention Heads Pruning}\quad Intra-attention pruning allows larger structure search space and more flexible models, but is intra-attention pruning more effective compared to attention heads pruning in practice? To answer the question, we conduct comparative attention heads pruning experiments.

We follow the GRAIN procedure, except for setting the pruning units to be attention heads and FFN hidden dimensions. The structure regularization strength is set to $0$, and the target model density is set to 5\%. Since each attention head has more parameters than each FFN hidden dimension, the importance scores of attention heads and FFN hidden dimensions are not directly comparable, so attention heads and FFN hidden dimensions can not be globally sorted and pruned.\footnote{A possible solution is to manually rescale the scores of different units. However, this also introduces additional hyperparameters.}  Hence, we sort and prune the two kinds of units independently and we have the freedom to set their densities as long as the model density is fixed to $5\%$. We experiment with five groups of $(\textrm{FFN}, \textrm{Heads})$ density,\footnote{ FFN (heads) density is defined as the percentage of the remained parameters in all FFNs (heads).} and the results are shown in Table \ref{table:headspruning}. \textbf{Intra+FFN} denotes pruning with intra-attention units. \textbf{Heads+FFN} denotes pruning with attention heads. {Heads+FFN} reaches its best performance when its ({FFN, Heads}) density is close to the (FFN, Heads) density of {Intra+FFN}, but {Intra+FFN} still outperforms {Heads+FFN} at different (FFN, Heads) densities. The results imply that 
intra-attention pruning is more effective than attention heads pruning.

\noindent\textbf{Model Structures}\quad
As we stated previously, intra-attention pruning tends to yield fragmented structures, which hinder running efficiency. We apply structure regularization (StructReg) to encourage generating models with less fragmented units. To get an intuitive understanding, Figure \ref{figure:structure} shows the structures of the models pruned with and without StructReg at model density $5\%$ on QNLI.\footnote{Structures of models on different tasks are listed in Appendix \ref{sec:structure}.} 
We first notice that with intra-attention pruning, attention heads take more diverse structures since the number of query and value units can differ.
The model pruned without StructReg holds 95 attention heads, where most heads contain only a few query or value units. The average query and value units per head are 9.8 and 8.2, respectively.
With StructReg, the model holds only 25 attention heads, and the average numbers of query and value units per head are 28.6 and 28.5. The number of heads is significantly reduced. We also find FFN layers are more severely pruned than attention heads, consistent with results in \citet{xia-etal-2022-structured}.

\noindent\textbf{Speed and Performance}\quad We next study the impacts of StructReg on speed and performance. We evaluate the latency with batch size 128 and sequence length 512 on an NVIDIA M40 GPU for all tasks. The results are shown in Figure \ref{figure:sr}. The latency of BERT$_{\textrm{base}}$ is around $3840$ms, far beyond the plots' range. The pruned models without StructReg only achieve about $4\times$ speedup. As the regularization strength $\alpha$ increases from $0$ to $0.3$, the latency decreases monotonically. At $\alpha=0.3$ (the leftmost marker in each plot),  models achieve $6\sim7\times$ speedups, notably faster than the unregularized ones.
The task performance is also affected by StructReg. As $\alpha$ increases from 0 to 0.3, the QNLI accuracy drops by 0.6\%, while SQuAD F1 increases by 0.4\%. There is no uniform trend in performance across different tasks. Nevertheless, compared to the gains in speedups, the variances in performance are marginal. 

\section{Conclusion}
This paper proposes GRAIN, a gradient-based structured pruning method that expands the structure search space by pruning with intra-attention structures. We provide a structure regularization strategy that encourages finding regular structures and helps achieve lower latencies. We also combine pruning with distillation. We propose to separate the gradients from different losses to reduce the interference. GRAIN is computationally efficient since it does not require pre-training or data augmentation. Experiments show that GRAIN achieves impressive high performance and outperforms other methods at different model densities on various natural language understanding tasks and meanwhile maintains competitive speedups.

\section*{Limitations}
\noindent\textbf{Inference Speed}\quad
At the same model size, the latencies of GRAIN on different tasks are relatively large compared to the methods like CoFi and TinyBERT. This is because GRAIN generates models with different head size, and the computation of these heads are not parallelized. Thus the resulting models are slower than the models with uniform attention structures. This problem could be relieved by introducing model structure regularization at a higher level or by some engineering techniques, such as merging heads with the same or similar size into a large matrix to increase parallelism.

\noindent\textbf{Backbone Models}\quad
GRAIN is designed for transformer-based models. Although the transformer is one of the most popular building blocks of NLP models, there are many other promising structures.  The effectiveness of GRAIN on model compression is possibly correlated with hardware lottery or software lottery \cite{hooker2020hardware}. In addition, we have only tested our method with the standard multi-head attention mechanism. Transplanting GRAIN to other attention mechanisms is possible, but the effectiveness has yet to be tested.

\section*{Acknowledgements}
This work is supported by the National Key Research and Development Program of China (Grant No. 2022YFC3303504).

\bibliography{custom}
\bibliographystyle{acl_natbib}

\clearpage
\appendix
\section{Reproducibility and Training Costs}
\label{sec:hyperparameters}

\textbf{Hyperparameters}\quad
We summarize the hyperparameters of our experiments in Table \ref{table:hyperparameters}. We use AdamW optimizer \cite{adamW}. The learning rate is scheduled with 10\% warm-up steps followed by a linear decay.

\noindent\textbf{Training Environment}\quad
All the training experiments are conducted on a single NVIDIA V100 GPU. The PyTorch \cite{DBLP:conf/nips/PaszkeGMLBCKLGA19} version is 1.8.1, the CUDA version is 10.2, and Transformers \cite{wolf-etal-2020-transformers} version is 4.10.0.

\noindent\textbf{Training Costs}\quad
It takes about 15 hours to finish training on MNLI and QQP, 11 hours on SQuAD, 5 hours on QNLI, 3 hours on SST-2, and 1 hour on CoNLL 2003.

\begin{table}[h]
\centering

\begin{tabular}{@{}lr@{}}
\toprule
\textbf{Hyperparameter}            & \textbf{Value}    \\ \midrule
\multirow{3}{*}{peak learning rate} & 3e-5 (GLUE) \\
                                    & 3e-5 (SQuAD) \\
                                    & 1e-4 (CoNLL 2003)  \\ \arrayrulecolor{black!50}\midrule
\multirow{3}{*}{number of epochs}   & 20 (GLUE)   \\
                                    & 20 (SQuAD)    \\
                                    & 40 (CoNLL 2003)    \\ \arrayrulecolor{black!50}\midrule
batch size                          & 32                 \\
temperature $\tau$                  & 8                  \\
start of pruning $p_s$              & 0.2                \\
end of pruning  $p_e$               & 0.4                \\
smoothing factor $\beta$            & 0.998              \\
regularization strength $\alpha$    & 0.3                \\
reduced embedding size $r$          & 192                \\ \arrayrulecolor{black!100}\bottomrule
\end{tabular}
\caption{Hyperparameters used in the experiments.}
\label{table:hyperparameters}
\end{table}

\begin{table}[t]
\centering

\begin{tabular}{@{}lccc@{}}
\toprule
\textbf{Task}  & \textbf{Train Size} & \textbf{Metric} & \textbf{\# Labels} \\ \midrule
\textit{English Task} & & & \\
QNLI          & 105k & Acc & 2  \\
MNLI          & 393k & Acc & 3  \\
QQP           & 364k & Acc & 2  \\
SST-2         & 67k  & Acc & 2  \\
SQuAD        & 88k  & F1  & N/A \\
CoNLL 2003   & 14k  & F1  & 9  \\ \midrule 
\textit{Chinese Task} & & & \\
OCNLI & 50k & Acc & 3 \\
TNEWS & 53k & Acc & 15 \\
CMRC 2018 & 10k 	& F1 & N/A \\
DRCD & 27k & F1 & N/A\\\bottomrule
\end{tabular}
\caption{Details of the datasets.}
\label{table:dataset}
\end{table}

\begin{table*}[th]
\centering
\small
\begin{tabular}{lcccccc@{}}
\toprule
\textbf{Datasets} & \textbf{MHA Layers} & \textbf{Total Heads} & \textbf{Query Units / Head} & \textbf{Value Units / Head} & \textbf{FFN Size} \\ \midrule
QNLI ($\alpha=0$)      & 12 & 95  & 9.8  & 8.2  & 87.9   \\
QNLI ($\alpha=0.3$)      & 12 & 25  & 28.6 & 28.5 & 106.1   \\ \midrule
MNLI ($\alpha=0$)       & 12 & 86  & 9.0  & 8.6  & 103.9   \\
MNLI ($\alpha=0.3$)       & 11 & 21  & 28.8 & 32.9 & 122.5   \\ \midrule
QQP ($\alpha=0$)        & 12 & 93  & 9.8  & 8.7  & 87.1   \\
QQP ($\alpha=0.3$)        & 12 & 26  & 27.5 & 26.4 & 113.5   \\ \midrule
SST-2 ($\alpha=0$)      & 12 & 101 & 4.2  & 8.9  & 120.2  \\
SST-2 ($\alpha=0.3$)      & 11 & 19  & 20.5 & 37.7 & 138.2   \\ \midrule
SQuAD ($\alpha=0$)      & 12 & 75  & 12.8 & 10.1 & 87.3    \\
SQuAD ($\alpha=0.3$)      & 12 & 23  & 33.0 & 30.8 & 108.0   \\ \midrule
CoNLL-03 ($\alpha=0$) & 12 & 91  & 6.1  & 9.1  & 114.5  \\
CoNLL-03 ($\alpha=0.3$) & 9  & 22  & 21.4 & 31.9 & 132.6  \\
 \bottomrule
\end{tabular}

\caption{Structures of the pruned models on different tasks at model density 5\%.}
\label{table:structure}
\end{table*}

\section{Dataset Statistics}\label{sec:dataset}
The details of the datasets are shown in Table \ref{table:dataset}.

\section{Structures of Pruned Models}\label{sec:structure}
Table \ref{table:structure} summarizes the structures of the pruned models on different tasks at model density 5\%. 

\section{Inference Speed vs. Performance}\label{sec:speed_results} 
\begin{figure*}[t!]
\centering
  \includegraphics[width=\linewidth]{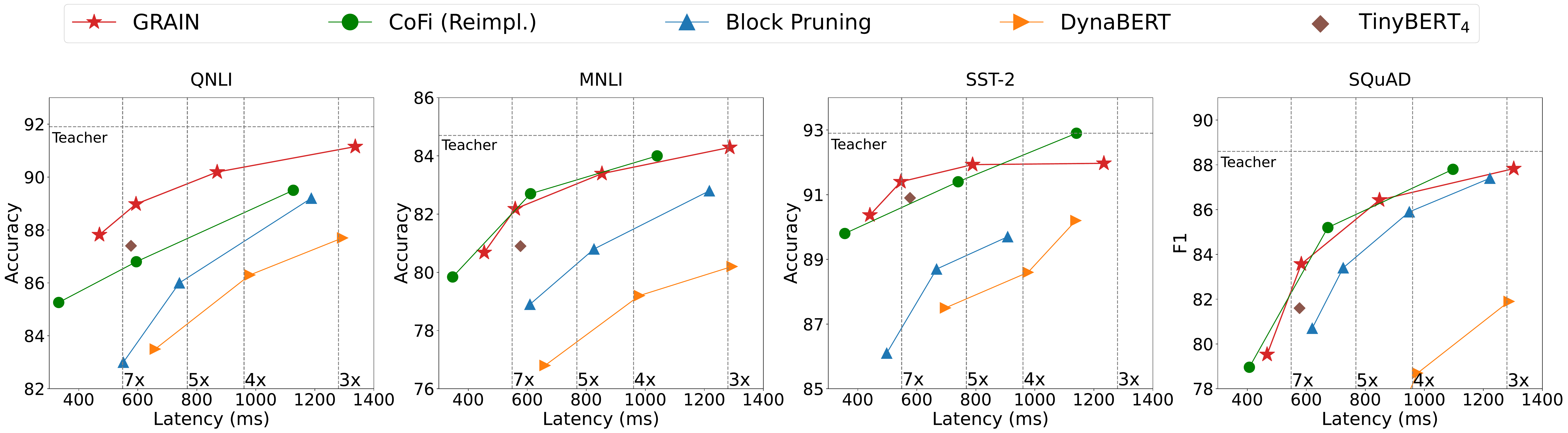}
  \caption{Performance vs. latency of GRAIN and other methods. The dashed vertical lines indicate the speedups relative to BERT$_\textrm{base}$.}
  \label{figure:main_results_speed}
\end{figure*}

Figure \ref{figure:main_results_speed} shows the latency of GRAIN and other methods on various tasks. All the measurements are conducted under the same environment (see the paragraph \textbf{Speed and Performance} in Section \ref{sec:analysis}). The structure regularization strength $\alpha$ is $0.3$. GRAIN achieves competitive speedups comparable to other methods.

\section{More Results}\label{app:results}
\subsection{Pruning RoBERTa}
We conduct GRAIN with RoBERTa-base \cite{liu2019roberta} on the same set of tasks and use the same hyperparameters as those in Table \ref{table:hyperparameters}. The results of GRAIN with BERT and RoBERTa at different model densities are shown in Table \ref{table:main_results_appendix}. The pruned RoBERTa outperforms pruned BERT at high densities, but at low densities, BERT surpasses RoBERTa on some tasks.

\subsection{Experiments on Chinese Tasks}

Due to the limited availability of results on model compression methods for Chinese tasks, we present the results of GRAIN on several Chinese tasks, providing a useful reference point for related works.

We evaluate GRAIN on the following Chinese tasks: OCNLI \cite{hu-etal-2020-ocnli}, an original Chinese natural language inference task; TNEWS \cite{clue}, a short text classification task for news; CMRC 2018 \cite{cmrc2018} and DRCD \cite{drcd}, two representative span-extraction Chinese machine reading comprehension tasks. The details of the datasets are shown in Table \ref{table:dataset}.

The learning rate is 1e-4 for CMRC 2018 and DRCD, 2e-5 for OCNLI and TNEWS; the number of epochs is 40 for CMRC 2018 and DRCD, 20 for OCNLI and TNEWS. Other hyperparameters are the same as those in Table \ref{table:hyperparameters}.
The teacher model is Chinese-RoBERTa-wwm-ext \cite{cui-etal-2021-pretrain}.

We report the mean score of 3 runs for each task using different random seeds. The results are shown in Table \ref{table:chinese_results}.

\begin{table*}[ht]
\centering
\small
\begin{tabular}{lcccccccc}
\toprule
\textbf{Model} &
  \begin{tabular}[c]{@{}c@{}}\textbf{QNLI}\\ {(Acc)}\end{tabular} &
  \begin{tabular}[c]{@{}c@{}}\textbf{MNLI}\\ {(m/mm Acc)}\end{tabular} &
  \begin{tabular}[c]{@{}c@{}}\textbf{QQP}\\ {(Acc)}\end{tabular} &
  \begin{tabular}[c]{@{}c@{}}\textbf{SST-2}\\ {(Acc)}\end{tabular} &
  \begin{tabular}[c]{@{}c@{}}\textbf{SQuAD}\\ {(F1 / EM)}\end{tabular} &
  \begin{tabular}[c]{@{}c@{}}\textbf{CoNLL-03}\\ {(F1)}\end{tabular} &
  \begin{tabular}[c]{@{}c@{}}\textbf{Model}\\ {\textbf{Size}}\end{tabular} &
  \begin{tabular}[c]{@{}c@{}}\textbf{Total}\\ {\textbf{Size}}\end{tabular} \\ \midrule
BERT$_{\textrm{base}}$ (teacher) & 91.9 & 84.7 / 85.0 & 91.2 & {92.9} & 88.6 / 81.1 & 91.2 & 85.1M        & 108.9M \\ 
RoBERTa$_{\textrm{base}}$ (teacher) & \textbf{93.0}  & \textbf{87.7 / 87.5}  & \textbf{91.7}  & \textbf{94.7}  & \textbf{91.5 / 84.9}  & \textbf{92.1}  & 85.1M        & 124.0M \\ \midrule
\rowcolor[HTML]{CBCBCB}
\textit{20\% Model Density}       &      &           &           &      &           &       &              &        \\
GRAIN                  & {91.2} & 84.3 / 84.2 & 91.0 & 92.0 & {87.8 / 79.9} & 90.4 & 17M (20\%)   & 23.4M  \\
GRAIN-R                  & \textbf{91.9} & \textbf{86.8 / 86.6} & \textbf{91.6} & \textbf{93.1} & \textbf{89.4 / 81.6} & \textbf{91.2} & 17M (20\%)   & 27.2M  \\ \midrule
\rowcolor[HTML]{D8D8D8}
\textit{10\% Model Density}       &      &           &           &      &           &       &              &        \\
GRAIN                       & 90.2 & {83.4 / 83.5} & 90.7 & 91.9 & 86.4 / \textbf{77.7} & 89.7 & 8.5M (10\%)   & 14.9M  \\
GRAIN-R                       & \textbf{90.9} & \textbf{\{85.0 / 85.0} & \textbf{91.0} & \textbf{92.2} & \textbf{86.5} / 77.6 & \textbf{90.7} & 8.5M (10\%)   & 18.7M  \\ \midrule
\rowcolor[HTML]{E4E4E4}
\textit{5\% Model Density}        &      &           &           &      &           &       &              &        \\
GRAIN                       & {89.0} & 82.2 / 82.5 & \textbf{90.4} & 91.4 & \textbf{83.6 / 73.7} & {88.3} & 4.3M (5.0\%) & 10.7M  \\ 
GRAIN-R                     & \textbf{89.4} & \textbf{83.1 / 83.0} & 90.3 & \textbf{91.6} & 82.4 / 71.9 & \textbf{89.7} & 4.3M (5.0\%) & 14.5M  \\ \midrule
\rowcolor[HTML]{F0F0F0}
\textit{3\% Model Density}                 &      &           &           &      &           &       &              &        \\
GRAIN                       & {87.8} & {80.7 / 81.1} & 90.0 & 90.4 & {79.5 / 68.4} & 86.8 & 2.6M (3.0\%) & 9.0M   \\ \bottomrule
\end{tabular}
\caption{Results of GRAIN (pruning BERT) and GRAIN-R (pruning RoBERTa) with model density varying from 3\% to 20\%.}
\label{table:main_results_appendix}
\end{table*}

\begin{table*}[ht]
\centering
\small
\begin{tabular}{lcccccc}
\toprule
\textbf{Model} &
  \begin{tabular}[c]{@{}c@{}}\textbf{OCNLI}\\ {(Acc)}\end{tabular} &
  \begin{tabular}[c]{@{}c@{}}\textbf{TNEWS}\\ {(Acc)}\end{tabular} &
  \begin{tabular}[c]{@{}c@{}}\textbf{CMRC 2018}\\ {(F1/EM)}\end{tabular} &
  \begin{tabular}[c]{@{}c@{}}\textbf{DRCD}\\ {(F1/EM)}\end{tabular} &
  \begin{tabular}[c]{@{}c@{}}\textbf{Model}\\ {\textbf{Size}}\end{tabular} &
  \begin{tabular}[c]{@{}c@{}}\textbf{Total}\\ {\textbf{Size}}\end{tabular} \\ \midrule
RoBERTa-wwm-ext (teacher) & 77.1 & 57.8 & 87.3 / 67.7 & 94.5 / 89.1  & 85.1M        & 101.7M \\  \midrule
\rowcolor[HTML]{CBCBCB}
\textit{20\% Model Density}       &      &           &       &       &              &        \\
GRAIN         & 75.4 & 56.9 & 87.3 / 67.7 & 93.8 / 88.5 & 17M (20\%)   & 21.6M  \\ \midrule
\rowcolor[HTML]{D8D8D8}
\textit{10\% Model Density}       &      &           &       &       &              &        \\
GRAIN          & 73.3 & 56.2 & 85.8 / 65.3  & 92.6 / 86.7 & 8.5M (10\%)   & 13.1M  \\ \midrule
\rowcolor[HTML]{E4E4E4}
\textit{5\% Model Density}        &      &           &     &       &              &        \\
GRAIN          & 70.2 & 55.6 & 83.5 / 61.1  & 90.6 / 83.4  & 4.3M (5.0\%) & 8.9M  \\  \bottomrule
\end{tabular}
\caption{Results of GRAIN (pruning Chinese RoBERTa-wwm-ext) on the development sets of  Chinese text classification and machine reading comprehension tasks.}
\label{table:chinese_results}
\end{table*}

\end{document}